\pdfoutput=1

\documentclass[11pt]{article}

\usepackage{ACL2023}

\usepackage{times}
\usepackage{latexsym}
\usepackage{graphicx}
\usepackage{float}
\usepackage{stfloats}

\usepackage[T1]{fontenc}

\usepackage[utf8]{inputenc}

\usepackage{microtype}

\usepackage{inconsolata}

\usepackage{tabularx} 
\usepackage{booktabs} 
\usepackage{array}    
\usepackage[table]{xcolor} 

\title{Biomedical Literature Q\&A System Using Retrieval-Augmented Generation (RAG)}

\author{
\normalfont
  \begin{tabular}{ccc}
    \normalfont Mansi Garg & \normalfont Lee-Chi Wang & \normalfont Bhavesh Ghanchi \\ 
    \normalfont \texttt{ghanchi@usc.edu} & \normalfont \texttt{leechiwa@usc.edu} & \normalfont \texttt{mansigar@usc.edu} \\
    \\[0.5em]  
    \normalfont Sanjana Dumpala & \normalfont Shreyash Kakde & \normalfont Yen Chih Chen \\
    \normalfont \texttt{dumpala@usc.edu} & \normalfont \texttt{skakde@usc.edu} & \normalfont \texttt{yenchihc@usc.edu} \\
  \end{tabular}
}

\begin{document}
\maketitle
\begin{abstract}
This work presents a Biomedical Literature Question Answering (Q\&A) system based on a Retrieval-Augmented Generation (RAG) architecture, designed to improve access to accurate, evidence-based medical information. Addressing the shortcomings of conventional health search engines and the lag in public access to biomedical research, the system integrates diverse sources including PubMed articles, curated Q\&A datasets, and medical encyclopedias to retrieve relevant information and generate concise, context-aware responses. The retrieval pipeline uses MiniLM-based semantic embeddings and FAISS vector search, while answer generation is performed by a fine-tuned Mistral-7B-v0.3 language model optimized using QLoRA for efficient, low-resource training. The system supports both general medical queries and domain-specific tasks, with a focused evaluation on breast cancer literature demonstrating the value of domain-aligned retrieval. Empirical results, measured using BERTScore (F1), show substantial improvements in factual consistency and semantic relevance compared to baseline models. The findings underscore the potential of RAG-enhanced language models to bridge the gap between complex biomedical literature and accessible public health knowledge, paving the way for future work on multilingual adaptation, privacy-preserving inference, and personalized medical AI systems.
\end{abstract}

\section{Introduction}

The proliferation of digital health resources has not eliminated the fundamental challenge of accessing accurate and trustworthy medical information. While general-purpose search engines provide broad accessibility, they frequently surface content that is incomplete, misleading, or sensationalized. Symptom-based queries, for example, often yield worst-case scenarios, fostering unnecessary anxiety and misinformation. In parallel, the gap between biomedical research publication and its translation into public knowledge persists, limiting the timely integration of scientific advances into clinical and personal decision-making processes.

Traditional information retrieval systems typically rely on rudimentary keyword-matching algorithms that fail to interpret the semantic and contextual subtleties of medical queries. Meanwhile, large language models (LLMs), despite their advancements in natural language generation, are prone to factual hallucinations, particularly when operating without explicit grounding in verified domain-specific knowledge. This dual limitation—uninformed retrieval and ungrounded generation—underscores a critical shortfall in existing medical Q\&A systems, especially when applied in high-stakes domains such as clinical decision support or patient education.

To address these limitations, this work explores the development of a Biomedical Literature Question Answering system capable of handling both general medical questions and domain-specific questions, leveraging a Retrieval-Augmented Generation (RAG) architecture.  RAG architectures combine neural retrieval mechanisms with generative models, enabling more accurate, contextually grounded answers through external knowledge integration. The proposed system focuses on integrating curated biomedical corpus with generative models to produce responses that are not only fluent but also empirically supported.

Data sources include publicly available medical datasets (e.g., Hugging Face, Kaggle), peer-reviewed articles from PubMed, and authoritative medical encyclopedias. Following preprocessing including document normalization, sentence-aware chunking, and noise removal—semantic embeddings were generated using the multi-qa-MiniLM-L6-cos-v1 model, a transformer variant fine-tuned for retrieval-centric tasks. The embeddings were indexed using the FAISS library to enable efficient top-k semantic similarity search.

The system was evaluated under three configurations: a baseline using vanilla Mistral-7B-v0.3, a retrieval-augmented Mistral-7B-v0.3 pipeline (RAG), and a retrieval-augmented Mistral-7B-v0.3 model further fine-tuned with QLoRA for parameter-efficient training. Evaluation was conducted using BERTScore (F1) to assess semantic fidelity relative to expert-annotated reference answers.

Results demonstrate that both retrieval augmentation and domain-specific corpus alignment significantly enhance response quality. Notably, experiments incorporating a breast cancer–specific corpus yielded the highest improvements, supporting the hypothesis that contextual relevance and biomedical grounding are crucial for reliable QA in specialized domains. This study thus contributes to the growing body of work on domain-adapted generative systems and highlights the potential of RAG-based architectures in bridging the gap between biomedical literature and accessible, trustworthy health information.

\section{Related Work}
Recent research has demonstrated the growing importance of Retrieval-Augmented Generation (RAG) frameworks in medical question-answering systems, particularly as a solution to the factual inconsistency and hallucination issues commonly observed in generative language models. Bora and Cuayáhuitl \citep{bora2024systematic} conducted a comprehensive comparative analysis involving open-source LLMs such as Flan-T5-Large, LLaMA-2-7B, and Mistral-7B, showing that the integration of RAG with fine-tuned Mistral-7B yielded the highest exact match accuracy (57\%) among the evaluated systems. This study reinforces the effectiveness of RAG in enhancing answer precision within biomedical domains.

Patel et al. \citep{patel2024virtual} introduced a virtual diagnosis chatbot built on LLaMA3, fine-tuned using the MedPalm dataset and extended with SigLIP to handle multimodal inputs. Their system achieved an 88\% accuracy in diagnostic tasks, highlighting the value of multimodal capabilities and structured data ingestion. The study further outlines future plans to integrate RAG, acknowledging its importance for handling text-based medical literature.

Sree et al. \citep{sree2024medgpt} proposed MEDGPT, a modular chatbot architecture that retrieves and synthesizes biomedical information from structured sources like PubMed and PDFs. The system showed strong performance across both physical and mental health diagnostics, demonstrating the utility of a hybrid RAG + agent-based design for maintaining contextually accurate and empathetic interaction. Compared to GPT-3.5 and LLaMA-2, their solution reported better latency, contextual relevance, and factual accuracy.

Zhao et al. \citep{zhao2024chatcad} presented ChatCAD+, a dual-module system that blends LLMs with hierarchical in-context learning for image-based diagnostic report generation and interactive patient consultations. This framework incorporates retrieval from trusted corpora such as the Merck Manual, providing robust clinical accuracy across diagnostic domains and demonstrating scalability in clinical settings.

In a broader review, Laymouna et al. \citep{laymouna2024chatbots} analyzed the role of chatbots in healthcare, identifying RAG as a key factor in enhancing the credibility and informativeness of health-related responses. Their findings underscore the need to ground chatbot output in curated biomedical knowledge to minimize misinformation and improve user trust.

Qiu et al. \citep{qiu2023large} provided a comprehensive survey of large AI models in health informatics, highlighting the integration of retrieval mechanisms with generative models as a promising strategy to address factuality challenges in medical AI applications. The authors discussed models such as ChatCAD+ that incorporate retrieval systems to enhance the factual grounding of generated outputs, underscoring the potential of combining large pretrained models with information retrieval from validated medical sources. Their work emphasizes the importance of retrieval-augmented approaches to improve explainability, transparency, and trustworthiness in AI-driven healthcare solutions.

Finally, Rajani and Ruparel \citep{rajani2023deep} explored a deep learning-based chatbot model for symptom classification and treatment recommendation. By integrating NLU components with neural classifiers and treatment mapping, the system automates end-to-end diagnosis workflows. Although not RAG-based, their modular framework highlights the significance of structured pipeline design for real-time clinical support and serves as a complementary baseline for evaluating retrieval-augmented alternatives.

 Collectively, these studies confirm the critical role of RAG in building accurate, trustworthy, and context-aware biomedical assistant systems, and provide a strong foundation for addressing both general medical questions and domain-specific challenges demonstrated in this work through evaluations on broad medical queries as well as a focused case study on breast cancer.

\section{Methodology}

To construct an effective biomedical question-answer system, three modeling strategies were explored and compared, each progressively enhancing contextual grounding and domain specialization. Successive approaches built upon previous iterations by incorporating domain knowledge and retrieval capabilities.

\subsection{Approaches}

\subsubsection{Approach 1: Vanilla Mistral-7B}

The baseline strategy utilized the Mistral-7B language model in a zero-shot configuration. Multiple variants were assessed, including \texttt{Mistral-7B-v0.1}, \texttt{Mistral-7B-v0.3}, and \texttt{Mistral-7B-Instruct}. Empirical evaluation across diverse medical queries indicated that \texttt{Mistral-7B-v0.3} consistently demonstrated superior factual consistency and coherence, establishing it as the default baseline. However, all vanilla Mistral-7B variants exhibited a notable limitation: the lack of access to domain-based biomedical sources. This constraint frequently resulted in hallucinated or shallow responses, highlighting the necessity of retrieval-augmented strategies.

\subsubsection{Approach 2: RAG with Mistral-7B-v0.3}

To address factual grounding limitations, a RAG pipeline was integrated into the system architecture. The implementation consisted of the following components:

\begin{itemize}
    \item Question embeddings were generated using the \texttt{multi-qa-MiniLM-L6-cos-v1} model, optimized for semantic search in question answering tasks.
    \item Relevant document chunks were retrieved from a domain-specific corpus using FAISS, a dense vector similarity search library.
    \item The top-\textit{k} semantically similar chunks were concatenated to the input prompt for \texttt{Mistral-7B-v0.3} (threshold value of k was chosen as 5), facilitating context-aware response generation.
\end{itemize}

The knowledge base for RAG was compiled from diverse, high-quality biomedical sources, including:

\begin{itemize}
    \item The BioASQ challenge dataset, a benchmark corpus for biomedical QA
    \item Curated web content from authoritative medical websites
    \item The Gale Encyclopedia of Medicine
    \item Peer-reviewed medical textbooks and StatPearls articles
\end{itemize}

\subsubsection{Approach 3: RAG with Fine-tuned Mistral-7B-v0.3 (QLoRA)}

Extending the RAG framework, the \texttt{Mistral{-}7B{-}v0.3} model was fine-tuned using domain-specific question–answer pairs from the MedQuAD dataset. The fine-tuning process employed QLoRA, a parameter-efficient approach enabling low-bit precision training suitable for large language models on resource-constrained hardware.

The fine-tuned model demonstrated the following enhancements:

\begin{itemize}
    \item Improved relevance and completeness of responses to biomedical queries
    \item More accurate usage of biomedical terminology, indicating effective domain adaptation
    \item Higher BERTScore-F1 performance, reflecting better alignment with expert-annotated reference answers
\end{itemize}

This optimized model was deployed within the existing RAG infrastructure, producing the most accurate and clinically aligned outputs among all evaluated configurations.

\subsection{RAG Pipeline}

The system architecture is structured around a Retrieval-Augmented Generation (RAG) pipeline, designed to combine the advantages of traditional information retrieval with the generative capabilities of large language models (LLMs). This hybrid approach enhances the factual accuracy, domain relevance, and contextual quality of generated answers. The RAG pipeline comprises the following interconnected stages as shown in Figure~\ref{fig:rag-pipe}:

\begin{figure}[htbp]
\centering
\includegraphics[width=0.45\textwidth]{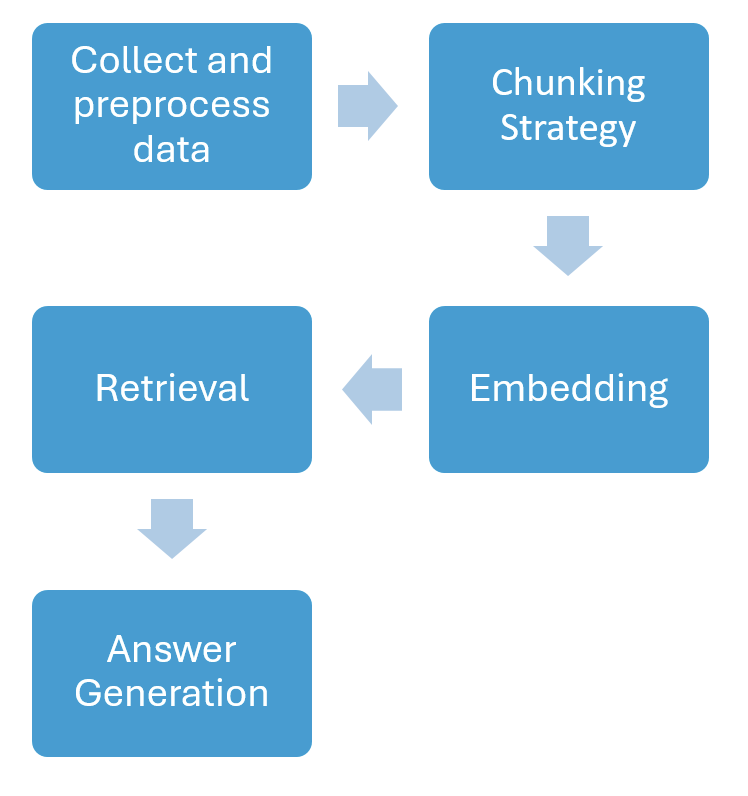}
\caption{RAG Pipeline}
\label{fig:rag-pipe}
\end{figure}

\subsubsection{Data Collection and Preprocessing}

Biomedical content was aggregated from publicly available sources, including Hugging Face datasets, Kaggle repositories, and scraped articles from PubMed and reputable medical encyclopedias. Preprocessing procedures involved converting various formats (e.g., PDF, HTML) into structured JSON, followed by the removal of non-informative content such as footnotes, references, and author details. Additional normalization steps were applied to clean special characters, redundant whitespace, and to standardize punctuation and text casing.

\subsubsection{Chunking Strategy}

To support effective retrieval, documents were segmented into semantically meaningful text blocks.

Multiple chunking strategies were evaluated:
\begin{itemize}
	\item Recursive Character Text Splitting: Hierarchical splitting with adjustable chunk size and overlap.
	\item Sentence/Paragraph-Aware Chunking: Division based on natural language boundaries to preserve contextual meaning.
	\item Dynamic/Adaptive Chunking: Modification of chunk size in response to logical or structural document features.
\end{itemize}

The objective was to preserve the integrity of medical content while ensuring compatibility with token constraints for embedding and retrieval; based on empirical results, Recursive Character Text Splitting was selected as the chunking method.

\subsubsection{Embedding}

Each chunk was encoded into a high-dimensional vector using the multi-qa-MiniLM-L6-cos-v1 model from Sentence Transformers. This model was selected due to its optimization for question-answering tasks and its superior balance between computational efficiency and semantic retrieval performance, in comparison with alternatives such as all-MiniLM-L6-v2 and SimCSE.

\subsubsection{Retrieval}

The resulting embeddings were indexed using Facebook AI Similarity Search (FAISS), enabling fast and scalable vector-based similarity searches at inference time. FAISS was configured for cosine similarity–based top-k (threshold value of k was chosen as 5) nearest neighbor retrieval to ensure semantic relevance. At runtime, the user query was embedded into the same vector space and matched against the FAISS index. The top-k relevant chunks were returned along with their similarity scores, forming the contextual input for the generative model. This retrieval mechanism grounded the generation step in factual, domain-specific biomedical content.

\subsubsection{Answer Generation}

\begin{figure}[htbp]
\centering
\includegraphics[width=0.5\textwidth]{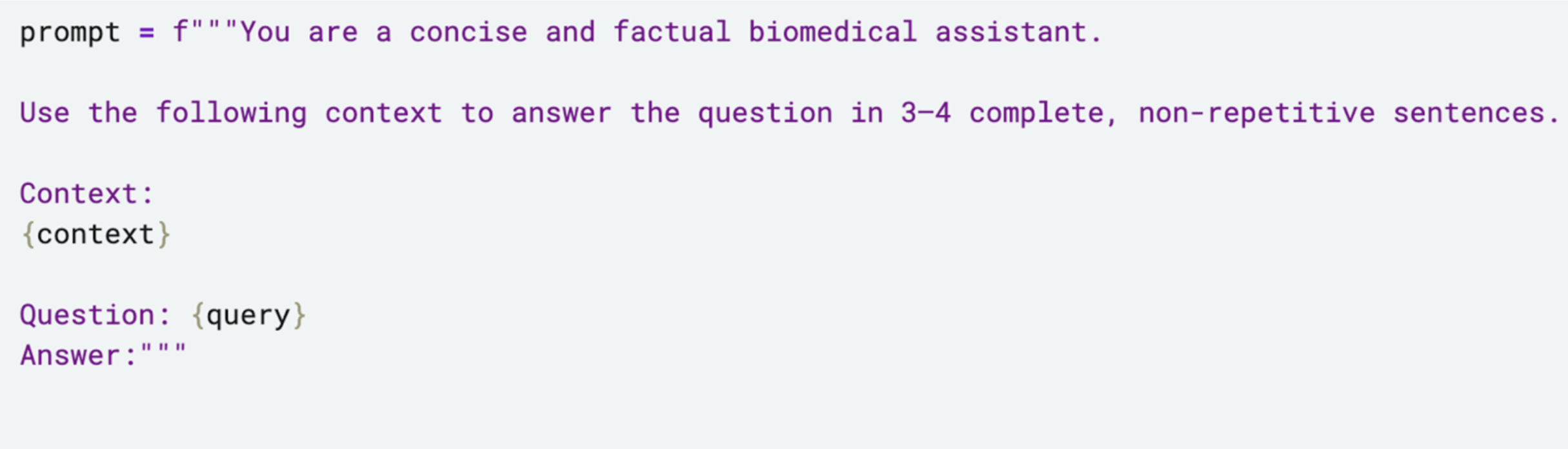}
\caption{Answer Generation Prompt Template}
\label{fig:prompt}
\end{figure}

Figure~\ref{fig:prompt} illustrates the prompt template used during answer generation. The prompt is engineered to instruct the language model to generate concise, medically accurate responses grounded in retrieved context.

\subsubsection{Prompt Components}
\begin{itemize}
	\item System Instruction:
“You are a concise and factual biomedical assistant.”
This sets the expected behavioral scope for the model.
	\item Explicit Guidance:
“Use the following context to answer the question in 3–4 complete, non-repetitive sentences.”
This enforces brevity and prevents redundancy in responses.
	\item Dynamic Insertion of Content:
{context} refers to retrieved text chunks from the FAISS index.
{query} refers to the input question provided by the user.
\end{itemize}

The final prompt includes both the context and the question, which are passed to the language model to produce a concise and grounded answer.

\subsubsection{Model Configurations}

The system was evaluated across three model variants:
\begin{itemize}
	\item Vanilla Mistral-7B-v0.3 (zero-shot; no retrieval augmentation)
	\item Mistral-7B-v0.3 with RAG (retrieval-augmented; no fine-tuning)
	\item Mistral-7B-v0.3 with RAG + fine-tuning (QLoRA fine-tuned on biomedical QA datasets)
\end{itemize}

The incorporation of fine-tuning with QLoRA resulted in notable improvements in factual consistency, contextual grounding, and response relevance, demonstrating the effectiveness of this configuration for biomedical question answering tasks.

\subsection{Fine-Tuning Strategy}
To enhance the alignment of the Mistral-7B-v0.3 model with biomedical question-answering tasks, a fine-tuning strategy was employed following the initial RAG pipeline setup.

\begin{itemize}
\item \textbf{Initial Approach – LoRA:} The first approach utilized Low-Rank Adaptation (LoRA), which introduces small trainable adapter weights into a frozen base model. Despite the application of 8-bit quantization, LoRA encountered memory and performance constraints, particularly due to the large parameter count of the 7B model.
\item \textbf{Transition to QLoRA:} To overcome these limitations, Quantized LoRA (QLoRA) was adopted. QLoRA enables training in 4-bit precision, significantly reducing memory requirements and allowing fine-tuning on a single A100 GPU without memory overflows. This setup supported scaling to larger datasets efficiently.

\item \textbf{Training Details:} Fine-tuning was conducted on the MedQuAD dataset, which comprises clinically validated medical Q\&A pairs. An instruction-tuned prompt format was used to encourage structured and concise responses from the model.

\item \textbf{Performance Improvements:} The fine-tuned model demonstrated:
\begin{itemize}
    \item Enhanced relevance and completeness of generated answers
    \item Improved utilization of biomedical terminology
    \item Higher BERTScore metrics compared to both zero-shot and retrieval-only variants
\end{itemize}

\item \textbf{Outcome:} Integrating the fine-tuned model into the RAG pipeline resulted in the most accurate and clinically grounded responses among all tested configurations.
\end{itemize}

\subsection{Final Working Design}

The Biomedical Literature Question Answering (Q\&A) system is implemented using a Retrieval-Augmented Generation (RAG) architecture. The pipeline begins with the acquisition and preprocessing of biomedical textual data sourced from publicly available datasets and web-scraped literature. These documents are segmented into semantically coherent chunks using advanced chunking techniques designed to preserve contextual integrity.

Each chunk is converted into a high-dimensional vector representation via a lightweight embedding model optimized for question answering. The resulting embeddings are indexed using FAISS (Facebook AI Similarity Search), a dense vector similarity search framework that facilitates efficient retrieval. At inference time, user queries are embedded in the same vector space and compared against the indexed corpus to retrieve the top-\textit{k} most relevant chunks.

The selected context is concatenated with the original user query and input to a fine-tuned \texttt{Mistral-7B-v0.3} language model. This configuration enables the generation of responses that are grounded in domain-specific biomedical evidence. Model performance is evaluated quantitatively using BERTScore-F1, which assesses both semantic alignment and factual accuracy relative to expert-annotated reference answers.

\begin{figure*}[htbp]
\centering
\includegraphics[width=\textwidth]{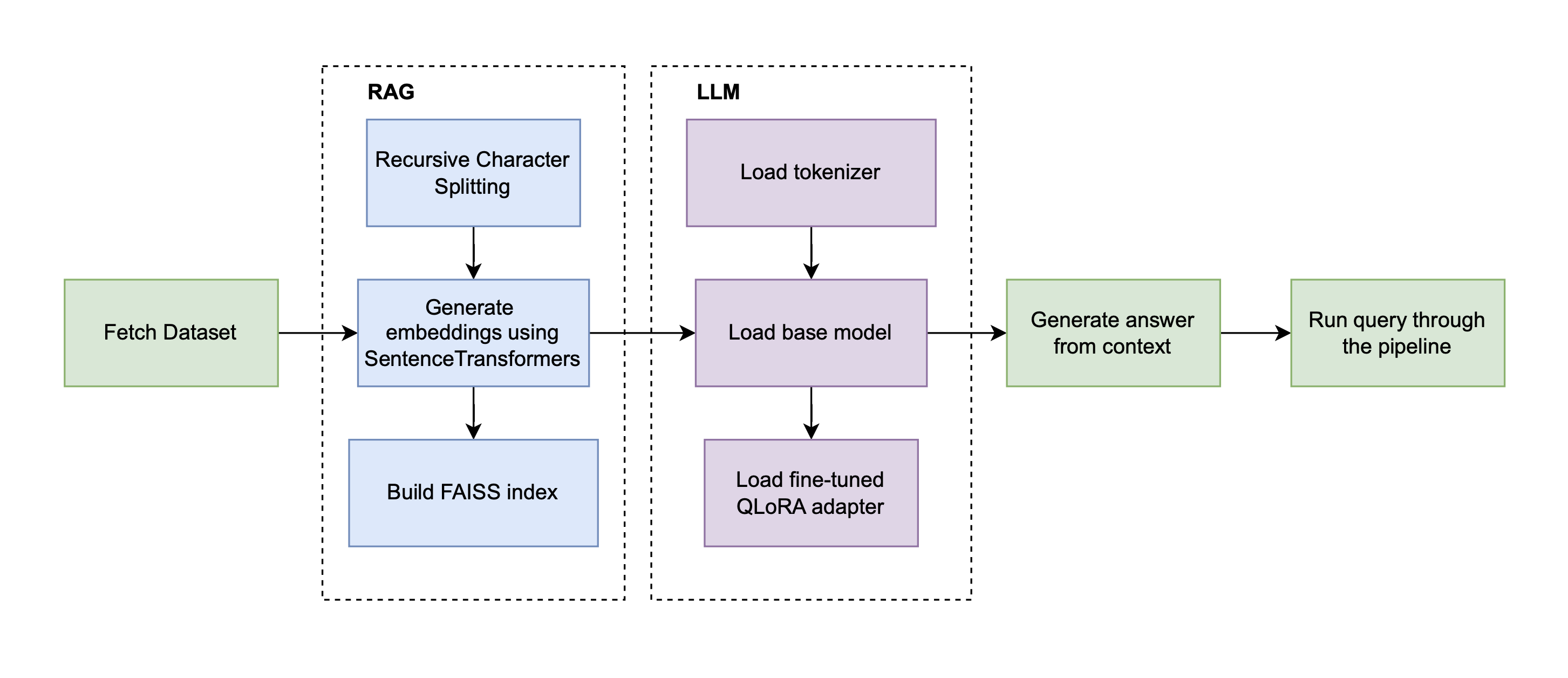}
\caption{Final Working Model}
\label{fig:citation-guide1}
\end{figure*}

The overall system architecture is illustrated in Figure~\ref{fig:citation-guide1} above.

\subsection{Domain-Specific Extension of RAG: Breast Cancer}

To evaluate the impact of domain-specific context on model performance, a specialized Retrieval-Augmented Generation (RAG) pipeline was constructed targeting the subdomain of breast cancer. Inspired by Halder et al. \citep{halder2024nlp}, who demonstrated the effectiveness of integrating structured health data into chatbot systems for patient prescreening, our approach similarly leverages curated domain-specific resources to enhance retrieval relevance. A corpus was curated by programmatically scraping peer-reviewed literature from PubMed, specifically filtering for articles related to breast cancer. The collected texts were subjected to sentence-aware chunking using the Natural Language Toolkit (NLTK) to maintain semantic coherence across context segments. Each chunk was then encoded into dense vector representations using the multi-qa-MiniLM-L6-cos-v1 model, a transformer architecture optimized for question–answer tasks and semantic retrieval.

These embeddings were indexed with FAISS (Facebook AI Similarity Search) to enable scalable and efficient similarity-based retrieval using cosine distance metrics. During inference, the system retrieved the top five most relevant document chunks for each input query from the breast cancer-specific index. These chunks served as contextual augmentation for answer generation using two configurations: the unmodified Mistral-7B-v0.3 language model and a RAG-enhanced variant. This setup facilitated a comparative analysis of generation quality under domain-targeted versus general-context retrieval conditions.

\subsection{Evaluation Metrics}

To evaluate the performance of our biomedical question answering system, we explored multiple evaluation strategies, both quantitative and qualitative, to determine the most suitable metric for capturing semantic and factual correctness in model-generated answers.

\subsubsection{Metric Exploration and Comparison}

Following evaluation metrics were experimented:

\begin{itemize}
    \item \textbf{Exact Match (EM):} Measures whether the predicted answer exactly matches the ground truth after normalization. While it provides a strict correctness signal, it fails to reward semantically correct but paraphrased answers.
    
    \item \textbf{BLEU Score:} Evaluates n-gram overlap between prediction and ground truth. Though useful for assessing language fluency, it is limited in capturing the deep semantic similarity required in biomedical contexts.
    
    \item \textbf{BERTScore:} Compares contextual embeddings of tokens using a pretrained model (\texttt{microsoft/deberta-xlarge-mnli} was used). It evaluates precision, recall, and F1 at the semantic level and is robust to paraphrasing.
\end{itemize}

After testing all of these, BERTScore was found to be the most informative. It captures both lexical and semantic similarity, accounts for paraphrasing, and provides detailed precision, recall, and F1 scores—making it more aligned with the needs of biomedical QA. Hence, BERTScore-F1 was selected as primary evaluation metric.

\subsubsection{Model and Prompt Combinations Tested}
To ensure a fair evaluation, several combinations of model versions and prompting strategies were tested. Each combination was evaluated using BERTScore on a curated set of 300 biomedical questions and reference answers. The results for each setup are presented in the Results section of this report.

Table~\ref{tab:model-eval-1} and Table~\ref{tab:model-eval-2} summarize the different setups we evaluated:

\begin{table}[htbp]
\centering
\footnotesize
\renewcommand{\arraystretch}{1.5}
\begin{tabularx}{\columnwidth}{|>{\raggedright\arraybackslash}X|>{\raggedright\arraybackslash}X|>{\raggedright\arraybackslash}X|}
\hline
\multicolumn{3}{|c|}{\textbf{Exploration with Generic Biomedical Dataset}} \\
\hline
\textbf{Approach} & \textbf{Model Version} & \textbf{Prompt Type} \\
\hline
Vanilla & Mistral-7B-v0.3 & \rule{0pt}{1.1em}QA-style \\
\hline
RAG & Mistral-7B-v0.3 & \rule{0pt}{1.1em}QA-style \\
\hline
RAG & Mistral-7B-v0.3 (QLoRA) & \rule{0pt}{1.1em}QA-style \\
\hline
\end{tabularx}
\caption{Model and Prompt Combinations Evaluated on Generic Biomedical Dataset}
\label{tab:model-eval-1}
\end{table}

\begin{table}[htbp]
\centering
\footnotesize
\renewcommand{\arraystretch}{1.5}
\begin{tabularx}{\columnwidth}{|>{\raggedright\arraybackslash}X|>{\raggedright\arraybackslash}X|>{\raggedright\arraybackslash}X|}
\hline
\multicolumn{3}{|c|}{\textbf{Exploration with Breast Cancer Dataset}} \\
\hline
\textbf{Approach} & \textbf{Model Version} & \textbf{Prompt Type} \\
\hline
Vanilla & Mistral-7B-v0.3 & \rule{0pt}{1.1em}QA-style \\
\hline
RAG & Mistral-7B-v0.3 & \rule{0pt}{1.1em}QA-style \\
\hline
RAG & Mistral-7B-v0.3 (QLoRA) & \rule{0pt}{1.1em}QA-style \\
\hline
\end{tabularx}
\caption{Model and Prompt Combinations Evaluated on Breast Cancer Dataset}
\label{tab:model-eval-2}
\end{table}

\subsection{User Interface}
\begin{figure}[htbp]
\centering
\includegraphics[width=0.5\textwidth]{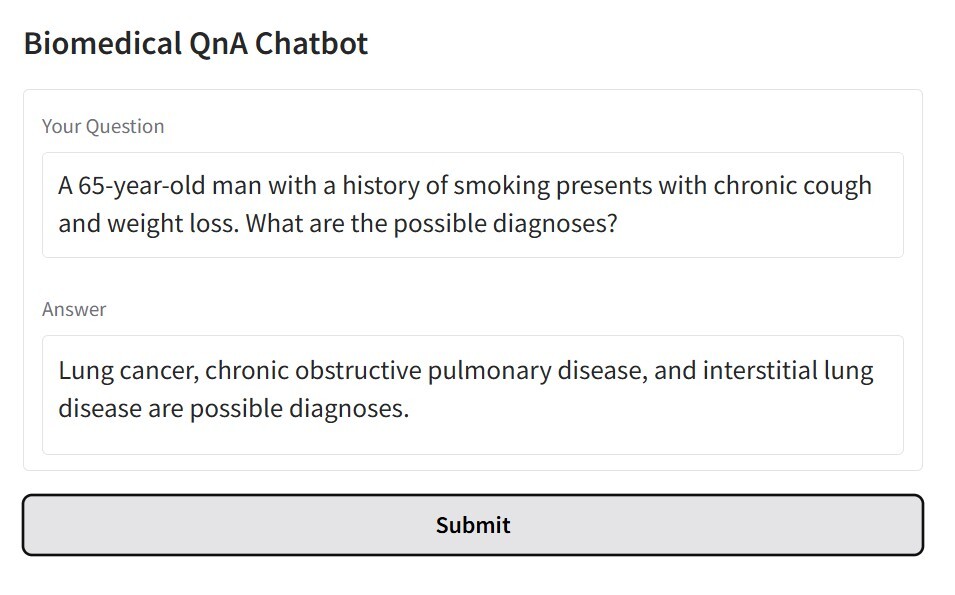}
\caption{Chatbot Interface}
\label{fig4ChatbotInterface}
\end{figure}

Our study initially adopted the approach used by Mohan et al.\citep{mohan2024medical}, implementing a Flask-based interface, but ultimately selected Gradio for its greater ease of deployment and user-friendly configuration. The web-based user interface enables users to input medical queries and receive concise, evidence-backed responses, integrating seamlessly with the RAG pipeline as shown in Figure \ref{fig4ChatbotInterface}.

\section{Evaluation and Results}
\subsection{General Medical Dataset Evaluation}
The system was first evaluated using the Comprehensive Medical Q\&A dataset, a publicly available benchmark known for its clinically sourced responses authored by medical professionals, including doctors, nurses, and pharmacists. The dataset is highly rated in terms of usability and credibility, with a score of 10.00 on Kaggle \citep{comprehensive_medical_qa}.

In contrast to the research done by Bhavani et al. \citep{bhavani2024chatbot}, which assessed chatbot performance using qualitative visualizations without employing standardized quantitative metrics, our study adopted BERTScore F1 as a formal evaluation measure. This choice ensures a more rigorous and replicable assessment of answer quality. As a semantic similarity measure, BERTScore captures both precision (the degree of relevance between the generated and reference answers) and recall (the completeness of the generated answer in covering reference content), thereby aligning closely with human evaluation standards \citep{zhang2019bertscore}. For instance, multiple correct paraphrases of a reference answer such as “How is asthma treated?” were assigned high BERT F1 scores, demonstrating the metric’s flexibility in recognizing synonymous medical expressions.

\begin{figure}[htbp]
\centering
\includegraphics[width=0.5\textwidth]{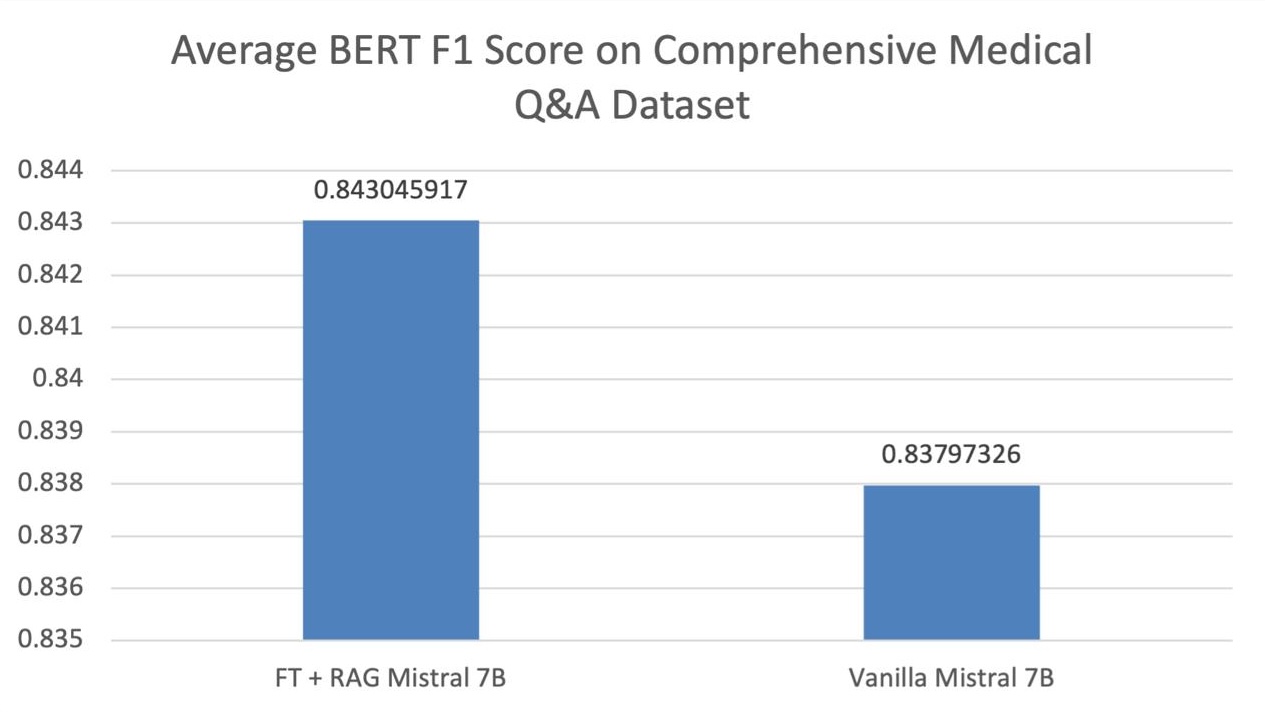}
\caption{Performance Comparison on Comprehensive Medical Q\&A Dataset}
\label{fig:med-comp}
\end{figure}

In terms of performance, the Fine-Tuned RAG Mistral-7B-v0.3 model achieved a BERT F1 score of 0.843, compared to 0.838 by the Vanilla Mistral-7B-v0.3 model, as shown in Figure~\ref{fig:med-comp}. Although the improvement was modest (approximately 0.605\%), such incremental gains are important in high-stakes medical domains where improved factual consistency can enhance trust and reliability.

To further assess the benefits of retrieval alignment, a custom evaluation dataset was constructed using the Gale Encyclopedia of Medicine. Prompts and reference answers were generated using GPT-4o to ensure that all test queries were semantically and topically consistent with the retrieval corpus. This approach facilitated controlled testing of the RAG system under domain-specific conditions.

\begin{figure}[htbp]
\centering
\includegraphics[width=0.47\textwidth]{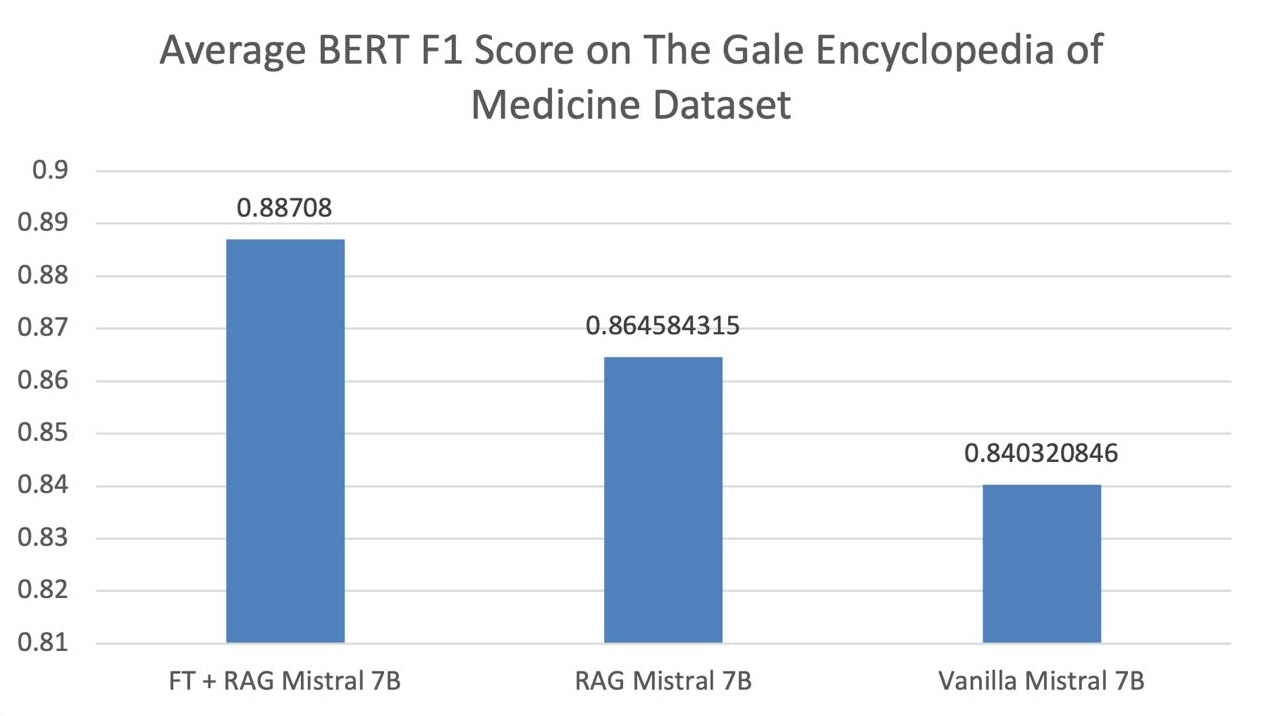}
\caption{Performance Comparison on Gale Encyclopedia-Based Custom Dataset}
\label{fig:gale}
\end{figure}

The results demonstrated that the Fine-Tuned RAG Mistral-7B-v0.3 model outperformed both its vanilla and RAG-only counterparts. Specifically, a 5.56\% improvement in BERTScore F1 was observed when comparing the fine-tuned RAG model to the baseline, as illustrated in Figure~\ref{fig:gale}.

These findings support the hypothesis that domain-aligned retrieval combined with fine-tuning substantially enhances model performance, particularly when the evaluation data is constructed to reflect the knowledge base used during inference. These experiments underscore the importance of high-quality, domain-specific data not only for training and retrieval but also for evaluation. The alignment between the model’s retrieval context and test set content played a pivotal role in uncovering the system’s full potential.

\subsection{Specific Medical Dataset (Breast Cancer) Evaluation}
\begin{figure}[ht]
\centering
\includegraphics[width=0.47\textwidth]{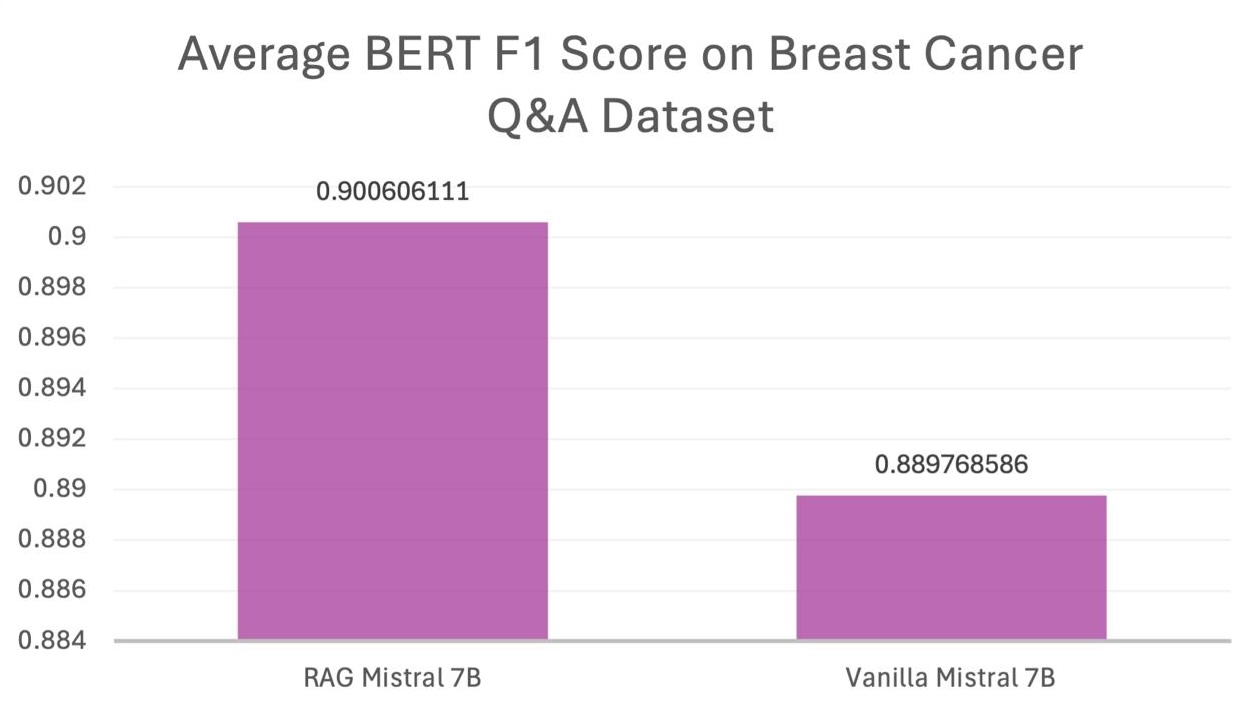}
\caption{Performance Comparison on Breast Cancer Scraped Dataset}
\label{fig:breast-cancer}
\end{figure}

Evaluation using BERTScore (F1) on the Breast Cancer dataset yielded two notable observations. First, the integration of breast cancer–specific documents into the retrieval pipeline led to a substantial improvement in performance relative to general medical corpora, with BERTScore F1 increasing from approximately 0.83–0.84 to 0.88–0.90 as shown in Figure~\ref{fig:breast-cancer}. 

\noindent

This outcome highlights the effectiveness of domain-specific context in enhancing the semantic relevance and factual accuracy of generated responses. Second, across all experimental conditions, models equipped with Retrieval-Augmented Generation (RAG) capabilities consistently outperformed their non-retrieval (vanilla) counterparts, even within the focused domain of breast oncology. These results underscore the importance of retrieval-based augmentation for improving response fidelity and grounding in biomedical question answering, particularly in domains where accuracy and specificity are critical.

\section{Conclusion}
This project presents a robust Retrieval-Augmented Generation (RAG) based biomedical Q\&A system that significantly enhances the accessibility and reliability of medical information. By integrating a hybrid retrieval strategy with fine-tuned large language models, the system effectively bridges the gap between complex biomedical literature and user-friendly, evidence-based answers. As illustrated in Table~\ref{tab:chatbot-qa}, the system provides concise and contextually grounded responses to various medical queries, demonstrating its practical utility. Through extensive experimentation, the fine-tuned RAG pipeline consistently outperformed baseline models, particularly in domain-specific contexts like breast cancer. The final system demonstrates notable improvements in factual accuracy, contextual relevance, and user trust. This work underscores the transformative potential of RAG architectures in healthcare applications, laying a foundation for future expansion into multilingual, privacy-preserving, and personalized medical AI systems.

\begin{table}[ht]
\centering
\footnotesize
\renewcommand{\arraystretch}{1.3}
\begin{tabularx}{\linewidth}{|>{\raggedright\arraybackslash}X|>{\raggedright\arraybackslash}X|}
\hline
\textbf{User Question} & \textbf{Chatbot Answer} \\
\hline
My mother is 58 years old and has been experiencing persistent fatigue, joint pain, and a facial rash that worsens in sunlight. What do these signs mean? & These symptoms could be a sign of a serious autoimmune condition called lupus. It is important for your mother to see a healthcare provider as soon as possible for evaluation. \\
\hline
How are patients monitored long-term after breast cancer remission? & The American Cancer Society recommends that patients who have had breast cancer should have a long-term follow-up plan to help monitor for possible cancer recurrence and to help the patient cope with the physical, psychological, and emotional issues of cancer survivorship. \\
\hline
Can breast cancer present without a lump? If so, what are the other signs? & Breast cancer can present with a variety of signs and symptoms. The most common sign is a lump or thickening in the breast, but it can also cause pain, a lump or thickening in the armpit, a change in the shape or size of the breast, or a nipple that is inverted, ulcerated or discharging. Other symptoms include a rash, skin irritation, itching, or nipple pain. \\
\hline
\end{tabularx}
\caption{Sample Questions and Responses from the Biomedical QnA Chatbot}
\label{tab:chatbot-qa}
\end{table}

\section{Limitations}
Despite the effectiveness of Retrieval-Augmented Generation (RAG) architectures, several limitations were encountered during development. The token limit of large language models imposes constraints on the context length, which can hinder comprehensive understanding in long-form clinical queries. Additionally, running these models locally poses challenges due to memory limitations, especially on machines without high-end GPUs. Fine-tuning large models also demands substantial computational resources and time, making it less accessible without dedicated infrastructure. On the data side, a notable limitation was that some PubMed articles lacked usable full text, reducing the richness of the retrievable knowledge base. Moreover, while FAISS is effective for dense vector search, it can miss relevant context if document chunking is not handled carefully. Overall, the effectiveness of RAG systems is strongly influenced by the quality of retrieval; when context is accurately and thoughtfully retrieved, it greatly enhances the relevance and reliability of the generated responses' retrieval quality.

\section{Future directions}
To further enhance the capabilities and societal impact of our biomedical Q\&A system, several future directions are envisioned. Currently, our chatbot aligns with the \textit{Preparation stage} of the Transtheoretical Model, as categorized in the systematic review by \citet{gentner2020review}, focusing on delivering tailored information and raising awareness to help users prepare for action. Building upon this stage, we aim to develop the system toward the \textit{Action} and \textit{Maintenance stages} by integrating interactive functionalities such as personalized recommendations, adaptive follow-up questions, and user-specific feedback mechanisms that encourage sustained engagement and informed decision-making.  

Another key area is the scaling of the medical corpus to incorporate a wider range of curated biomedical literature, multilingual datasets, and real-world clinical guidelines. This expansion would not only improve the model’s factual grounding but also enable broader applicability across specialties and patient populations. Additionally, as the system moves toward potential deployment in healthcare settings, ensuring privacy-preserving inference will be essential. Techniques such as differential privacy, secure multi-party computation, or encrypted model serving could help safeguard sensitive health information while maintaining inference efficiency.

A critical advancement involves improving evaluation methodologies beyond BERTScore by introducing clinically meaningful benchmarks and human-in-the-loop assessments involving medical professionals. The system could also benefit from context-aware interaction, allowing it to maintain a history of user queries and generate more coherent multi-turn dialogues. Personalization through a user profiling mechanism may further enhance relevance and clarity by adjusting responses based on user expertise (e.g., layperson vs. medical professional) or prior interactions.  

Lastly, the integration of multilingual support is a crucial step toward improving global accessibility, ensuring that accurate and reliable medical information is available to users across linguistic and cultural boundaries.


\end{document}